\documentclass{article}
\PassOptionsToPackage{numbers,sort&compress}{natbib}
\usepackage[final]{neurips_2025}

\usepackage{scalerel}
\usepackage{graphicx}
\usepackage{amsmath}
\usepackage{amssymb}
\usepackage{booktabs}

\usepackage{scalerel}  
\usepackage{epsfig}
\usepackage{graphicx}
\usepackage{amsmath}
\usepackage{pifont}
\usepackage{amssymb}
\usepackage{booktabs}
\usepackage{epsfig} 
\usepackage{makecell}
\usepackage{silence}
\usepackage{epstopdf}
\usepackage{color, colortbl}
\usepackage{float}  
\usepackage{caption}
\usepackage{multirow} 
\usepackage{url}
\usepackage[numbers]{natbib}
\usepackage{placeins}
\usepackage[group-separator={,}]{siunitx}
\usepackage[utf8]{inputenc}
\usepackage{quoting,xparse}
\usepackage{paralist}
\usepackage{listings}

\usepackage[accsupp]{axessibility}  

\usepackage{pdfrender}
\newcommand{\boldcheckmark}{%
  \textpdfrender{
    TextRenderingMode=FillStroke,
    LineWidth=0.7, 
  }{\checkmark}%
}

\newcommand{\boldxmark}{%
  \textpdfrender{
    TextRenderingMode=FillStroke,
    LineWidth=.5pt, 
  }{\xmark}%
}

\NewDocumentCommand{\bywhom}{m}{
  {\nobreak\hfill\penalty50\hskip1em\null\nobreak
   \hfill\mbox{\normalfont(#1)}%
   \parfillskip=0pt \finalhyphendemerits=0 \par}%
}

\NewDocumentEnvironment{pquotation}{m}
  {\begin{quoting}[
     indentfirst=true,
     leftmargin=\parindent,
     rightmargin=\parindent]\itshape}
  {\bywhom{#1}\end{quoting}}

\newcommand{\modelname}{AUGUSTUS\xspace}

\newcommand{\xmark}{\ding{55}}%

\newcommand{\comment}[1]{}

\definecolor{LightCyan}{rgb}{0.88,1,1}
\definecolor{Gray}{gray}{0.9}
\definecolor{Pink}{rgb}{1, 0, 1}
\definecolor{azure}{rgb}{0.0, 0.44, 1.0}
\definecolor{bleudefrance}{rgb}{0.19, 0.55, 0.91}
\definecolor{cobalt}{rgb}{0.0, 0.28, 0.67}
\definecolor{electricpurple}{rgb}{0.75, 0.0, 1.0}
\definecolor{cvprblue}{rgb}{0.21,0.49,0.74}
\definecolor{lightblue}{rgb}{0.85, 0.95, 1}
\definecolor{lightorange}{rgb}{1, 0.95, 0.85}
\definecolor{lightpink}{rgb}{1, 0.9, 0.95}
\usepackage[pagebackref,breaklinks,colorlinks,citecolor=violet]{hyperref}

\definecolor{LightCyan}{rgb}{0.88,1,1}

\usepackage[capitalize]{cleveref}
\crefname{section}{Sec.}{Secs.}
\Crefname{section}{Section}{Sections}
\Crefname{table}{Table}{Tables}
\crefname{table}{Tab.}{Tabs.}
\crefname{algocf}{Algo.}{algos.}
\Crefname{algocf}{Algorithm}{Algorithms}
\usepackage[dvipsnames]{xcolor}

\usepackage[ruled,linesnumbered]{algorithm2e}
\usepackage{color} 

\definecolor{codegreen}{rgb}{0,0.6,0}
\definecolor{codegray}{rgb}{0.5,0.5,0.5}
\definecolor{codepurple}{rgb}{0.58,0,0.82}
\definecolor{backcolour}{rgb}{0.95,0.95,0.92}
\definecolor{darkgreen}{rgb}{0.0, 0.5, 0.0}

\newcommand{\tcheck}{\color{darkgreen}\boldcheckmark}
\newcommand{\tx}{\color{red}\boldxmark}
\newcommand{\tna}{\color{gray}\textbf{N/A}}

\lstdefinestyle{mystyle}{
    backgroundcolor=\color{backcolour},   
    commentstyle=\color{codegreen},
    keywordstyle=\color{magenta},
    numberstyle=\tiny\color{codegray},
    stringstyle=\color{codepurple},
    basicstyle=\ttfamily\footnotesize,
    breakatwhitespace=false,         
    breaklines=true,                 
    captionpos=b,                    
    keepspaces=true,                 
    numbers=left,                    
    numbersep=5pt,                  
    showspaces=false,                
    showstringspaces=false,
    showtabs=false,                  
    tabsize=2
}

\usepackage{microtype}
\usepackage{url}
\usepackage{booktabs}

\usepackage{lineno}

\definecolor{darkblue}{rgb}{0, 0, 0.5}
\hypersetup{colorlinks=true, citecolor=darkblue, linkcolor=darkblue, urlcolor=darkblue}

\title{AUGUSTUS: An LLM-Driven Multimodal Agent System with Contextualized User Memory} 

\author{
  Jitesh Jain\textsuperscript{1}\thanks{Equal Contribution.\textsuperscript{$\dagger$}Work done during SM's time at Georgia Tech. \\ \textit{Note:} The research presented in this work was conducted in late 2023 to early 2024.} \quad
  Shubham Maheshwari\textsuperscript{2$*\dagger$} \quad
  Ning Yu\textsuperscript{3} \quad
  Wen-mei Hwu\textsuperscript{4} \quad
  Humphrey Shi\textsuperscript{1}\\[0.3em]
  {\textsuperscript{1}SHI Labs @ Georgia Tech \quad \textsuperscript{2}Adobe \quad \textsuperscript{3}Netflix Eyeline Studios \quad \textsuperscript{4}UIUC}
}

\begin{document}

\maketitle

\begin{center}
    \centering
    \captionsetup{type=figure}
    \vspace{-0.6cm}
    \includegraphics[width=0.95\textwidth]{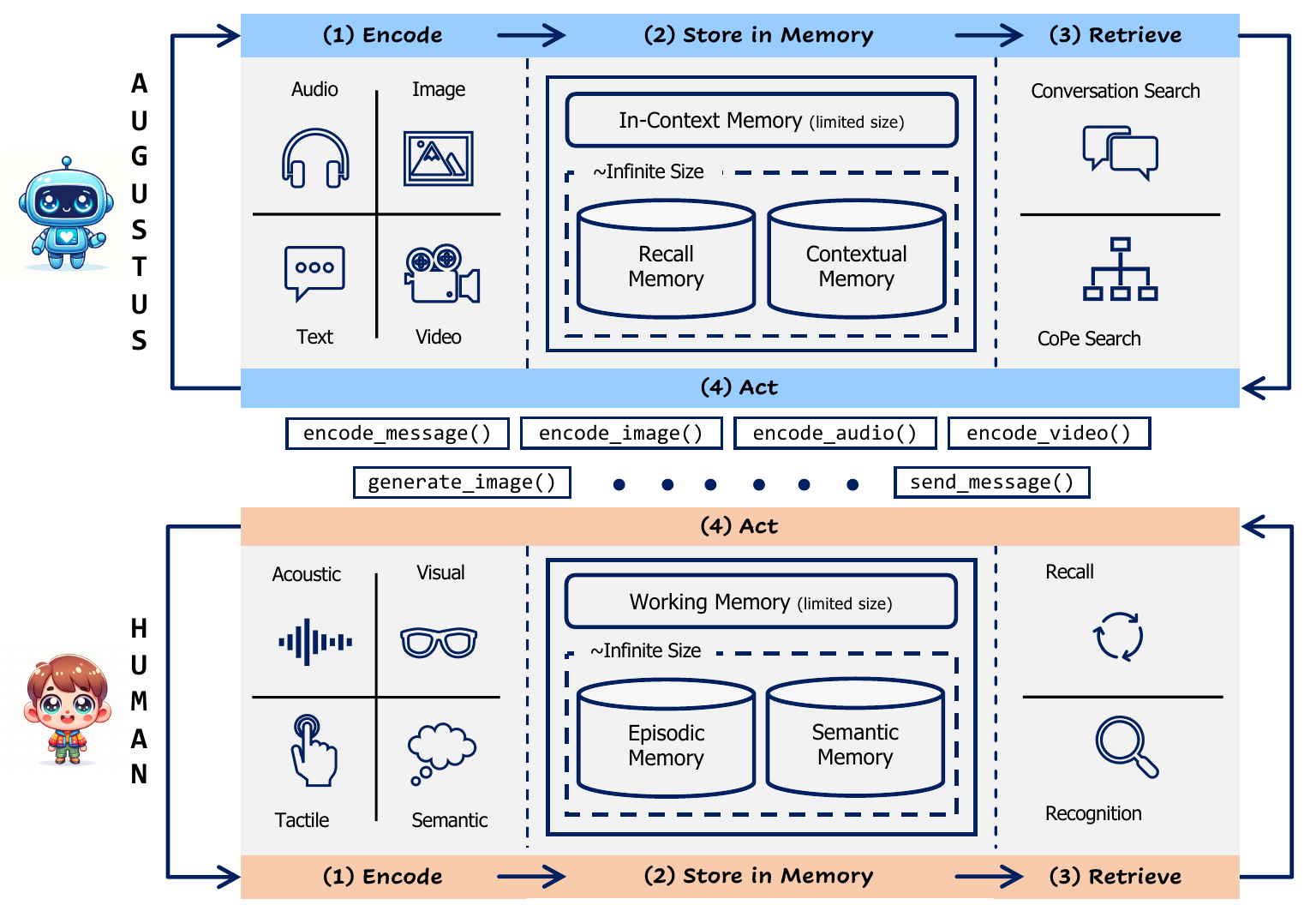}
    \caption{\textbf{Alignment with Human Cognition System.} \modelname first encodes information for storage into long-term (external) memory, using the limited context length in LLMs as working memory and retrieving information upon the need to act.}
    \label{fig:teaser}
\end{center}%

\begin{abstract}

Riding on the success of LLMs with retrieval augmented generation (RAG), there has been a growing interest in augmenting agent systems with external memory databases. However, the existing systems focus on storing text information in their memory, ignoring the importance of multimodal signals. Motivated by the multimodal nature of human memory, we present \textbf{\modelname}, a multimodal agent system aligned with the ideas of human memory in cognitive science. Technically, our system consists of 4 stages connected in a loop: (i) \textit{encode}: understanding the inputs; (ii) \textit{store in memory}: saving important information; (iii) \textit{retrieve}: searching for relevant context from memory; and (iv) \textit{act}: perform the task. Unlike existing systems that use sequential key-value databases, we propose conceptualizing information into semantic tags and associating the tags with their context to store them in a graph-structured multimodal contextual memory for efficient concept-driven retrieval. Our system outperforms the traditional multimodal RAG approach while being 3.5 times faster for ImageNet classification and outperforming MemGPT on the MSC benchmark.

\end{abstract}

\section{Introduction}
\label{sec:intro}

Processing information is a crucial part of a human's existence~\citep{Gazzaniga2018CognitiveNeuroscience}. At the core of processing lies the memory system~\citep{HarvardBokCenter2023} in humans, persistently storing, retrieving, and modifying multimodal information across dynamic working and long-term memory. The human ability to react to every situation based on the context is guided by memory to a large extent~\citep{Schacter1997, Kandel2000}. Recently, LLMs have emerged as agent systems~\citep{surismenon2023vipergpt, yang2023mmreact, Park2023GenerativeAgents} that think~\citep{wei2023chainofthought, yao2023tot}. Although impressive, we find that the role of memory in these agent systems is relatively less studied, even less so for multimodal memory. We observe that most existing systems either maintain a vector database knowledge base, performing retrieval augmented generation (RAG) when necessary~\citep{openai2023gpt4, geminiteam2023gemini} or do not maintain or access any form of external memory, therefore, only capable of planning the execution of tools defined in their system prompt~\citep{Gupta2022VisProg, surismenon2023vipergpt}. More recently, a few works~\citep{packer2023memgpt, Park2023GenerativeAgents, wu2024oscopilot} proposed maintaining a dynamic long-term memory, which is updated as the agent interacts with other agents or the user. However, the mentioned systems (and their long-term memory) only support the text modality during their interactions with a user. Intuitively, an agent system should be able to retain all context from user conversations, including any relevant image, audio, or video data for future conversations. Therefore, developing an agent system that can process, store, and retrieve multimodal information is imperative~\citep{Baddeley2014}. 

Working towards such a system, we present \textbf{\modelname}, a multimodal agent storing user context in memory, seeking answers to the following questions:

\noindent
(i) \textbf{Is it possible to establish an analogy between existing agent systems and the human cognition system?} Based on past research in cognitive neuroscience, working memory is considered to have limited capacity~\citep{working_memory_limit}. In contrast, long-term memory is believed to have unlimited capacity~\citep{Cowan2008}. We recognize the similarity between the working memory in humans and the in-context memory in large language models (LLMs), with both having a limited size and being the primary source for planning/reasoning for making a decision~\citep{Cowan2014WorkingMemory_plan}. We observe another analogy between the cue-driven retrieval~\citep{Tulving1973} in humans from long-term memory and RAG~\citep{rag} using external memory in LLMs. Lastly, we focus on two main types of long-term memory~\citep{Tulving1972}: \textit{episodic memory}, responsible for storing temporally dated information, and \textit{semantic memory}, responsible for storing concepts and relation among the held concepts. Recently, Packer \emph{et al.}~\citep{packer2023memgpt} proposed using a \textit{recall memory} to store all user (text) conversations in chronological order. We realize the similarities between episodic and recall memory while developing \modelname. Moreover, we equip \modelname with a contextual memory to store conceptualized information, similar to semantic memory. We present more details on the analogy in \cref{sec:align_cog}.

\noindent
(ii) \textbf{Building upon the observed analogies, is it possible to develop a general multimodal agent system good at remembering context about the user?} Recently, there has been a growing interest in developing any-to-any systems~\citep{xu2023versatile, wu2023nextgpt, wang2024mllmtool, wang2024modaverse} due to their modular design. We believe a general system should be able to understand and generate multiple modalities. Therefore, we also leverage existing foundation models~\citep{jiang2024mixtral, whisper_captioning, sd, 2023i2vgenxl} as tools to equip \modelname to understand and generate data for multiple modalities. At the center of our system lies an LLM, the planner responsible for creating function calls to execute the corresponding tool. We also provide our system with the appropriate tools to store and retrieve information in/from two types of external (long-term) memory databases: recall~\citep{packer2023memgpt} and contextual memory. We attempt to imitate the humans' ability to conceptualize information~\citep{Tulving1972} for storage in semantic memory by equipping our agent with the ability to abstract information into semantic tags (concepts) to store in the contextual memory. The tags in the contextual memory are connected to context nodes that store information about the user from the corresponding conversation snapshot~\citep{Anderson1996}, including any media. Thus, the contextual memory stores the multimodal context about the user. Additionally, we introduce the Contextual-Personalized (CoPe) search mechanism to perform concept-driven retrieval of information from the contextual memory to provide personalized responses (\cref{fig:personalization}) to the user.

To summarize, our contributions are three-fold:

\begin{compactitem}
    \item We establish an analogy between human cognition and agent systems, introducing \modelname with a contextual memory aligned with the ideas of a multimodal memory system in humans from cognitive neuroscience.
    \item We equip our agent system with the ability to conceptualize information for storage in contextual memory for a more efficient memory system. Moreover, we introduce a two-stage CoPe search for effective concept-driven retrieval.
    \item We provide empirical evidence on the effectiveness of our system, outperforming the traditional multimodal RAG approach while being 3.5 times faster. Moreover, \modelname outperforms MemGPT~\citep{packer2023memgpt} on the MSC benchmark.
\end{compactitem}
\section{Related Works}
\label{sec:rel_work}

\subsection{LLM-Driven Agent Systems}

With the rise of LLMs for various language and multimodal tasks~\citep{liu2023llava,jiang2024mixtral}, the community has leveraged LLMs as planners~\citep{yao2023react, shen2023hugginggpt} in agent systems, having three modules: \textit{tool collection}, \textit{planning module}, and \textit{memory}. 

The \textit{tool collection} includes multimodal foundation models for understanding the user query~\citep{whisper_captioning, lin2023video, touvron2023llama2} and generating the response~\citep{sd, 2023i2vgenxl, fu2024mgie}. 

The \textit{planning module} is an LLM that generates a sequence of execution steps using its reasoning ability to fulfill the user query. The format and instructions for the step generation are passed as few-shot or zero-shot in-context prompt to the LLM to achieve successful step generation, which can either be a natural language~\citep{wei2023chainofthought, wu2023visual} or programs~\citep{surismenon2023vipergpt}, or both~\citep{packer2023memgpt, gao2023assistgpt}.

The community has incorporated long-term external memory databases into agent systems~\citep{wang2023jarvis1, wu2024oscopilot} to overcome the LLMs' limited context length. 

For \modelname, we also use off-the-shelf foundation models during the \textit{encode} and \textit{act} stages. For planning, inspired by the success and elegance of planning through function calling in MemGPT~\citep{packer2023memgpt}, we also follow the same strategy as a combination of natural language and programs. We use a fine-tuned version of Mixtral-8x7b~\citep{jiang2024mixtral, thebloke_dolphin_2024} as our planning module for improved function calling abilities. \modelname can directly access the contents in the external memory, making our system's memory management completely autonomous~\citep{packer2023memgpt}.

\begin{table}[t!]
\centering
\caption{\textbf{\modelname' Tool Collection.} We equip our system with various tools for smooth multimodal operation. Our experiments use tools in \textbf{bold} as default choices.}
\vspace{0.3em}
\scalebox{0.83}{
\begin{tabular}{l@{\hspace{3mm}}l@{\hspace{3mm}}l@{\hspace{3mm}}l}
\toprule
{Stage} & {Tool} & {Modality} & {Model Access} \\
\midrule
\multirow{5}{*}{{Encode}} &  \textbf{Video-LLaVA}~\citep{lin2023video} &  {image; video}  &  \color{darkgreen}{Open-Source}  \\
&  \textbf{WhisperX}~\citep{bain2022whisperx} &  {audio \textit{(speech)}}  &  \color{darkgreen}{Open-Source}  \\
&  \textbf{Whisper-Captioner}~\citep{whisper_captioning} &  {audio}  &  \color{darkgreen}{Open-Source}  \\
& \textbf{dolphin-2.6-mixtral-8x7b}~\citep{thebloke_dolphin_2024}  &  {text}  &  \color{darkgreen}{Open-Source} \\
&  GPT-4~\citep{openai2023gpt4} &  {text}  &  \color{red}{Close-Source}  \\
\midrule
\multirow{2}{*}{{Store in Memory}} & \textbf{Recall Memory (sqliteDB)} &  {text}  &  \color{gray}{Not Applicable} \\
 &  \textbf{Contextual Memory (arangoDB)} &  {image; video; audio; text}  &  \color{gray}{Not Applicable}  \\
\midrule
\multirow{7}{*}{{Generate}} &  \textbf{SD-2.1}~\citep{sd} &  {image}  &  \color{darkgreen}{Open-Source} \\
&  \textbf{MGIE}~\citep{fu2024mgie} &  {image \textit{(edit)}}  &  \color{darkgreen}{Open-Source} \\
&  LaVIT~\citep{jin2023unified} &  {image \textit{(edit)}}  &  \color{darkgreen}{Open-Source} \\
&  \textbf{MAGNeT}~\citep{ziv2024magnet} &  {audio}  &  \color{darkgreen}{Open-Source} \\
&  \textbf{I2VGen-XL}~\citep{2023i2vgenxl} &  {video}  &  \color{darkgreen}{Open-Source} \\
& \textbf{dolphin-2.6-mixtral-8x7b}~\citep{thebloke_dolphin_2024}  &  {text}  &  \color{darkgreen}{Open-Source} \\
&  GPT-4~\citep{openai2023gpt4} &  {text}  &  \color{red}{Close-Source}  \\
 

\bottomrule
\end{tabular}

}
\label{tab:model_card}
\end{table}

\subsection{Agent Systems with Memory}

Inspired by the success of retrieval-augmented-generation (RAG)~\citep{rag, asai2024selfrag}, existing agent systems have also incorporated an external memory in vector databases to overcome the limitations of limited context in LLMs. Park \textit{et al.}~\citep{Park2023GenerativeAgents} maintained agent-specific memory to store the experiences of each persona as different agents interacted with each other in a simulation. MemGPT~\citep{packer2023memgpt} proposed holding two types of external databases: storing raw conversation history in its recall memory and essential information about the user in archival memory. OS-Copilot~\citep{wu2024oscopilot} proposed a more cognitively aligned approach, storing the user and the agent's action history information in a declarative memory database while storing information about tools in a procedural memory database to support learning new tools for acting as an operating system. Although impressive, these systems only support storing text information in their external memory databases. More recently, JARVIS-1~\citep{wang2023jarvis1} proposed storing both the plan and images in memory to improve performance at solving tasks in the Minecraft game environment. However, JARVIS-1 maintained a non-hierarchical memory while only supporting image and text modalities, not aligning with the human memory system (\cref{sec:align_cog}), leading to inefficient retrieval with increased memory size. In contrast to the existing agent systems, our \modelname maintains the contextual memory as a hierarchical multimodal database, storing semantic concepts at the top level with the associated context stored under the corresponding concepts, reducing the search space.
\section{System Description}
\label{sec:system}

\begin{figure}[!t]
    \centering
    \includegraphics[width=\textwidth]{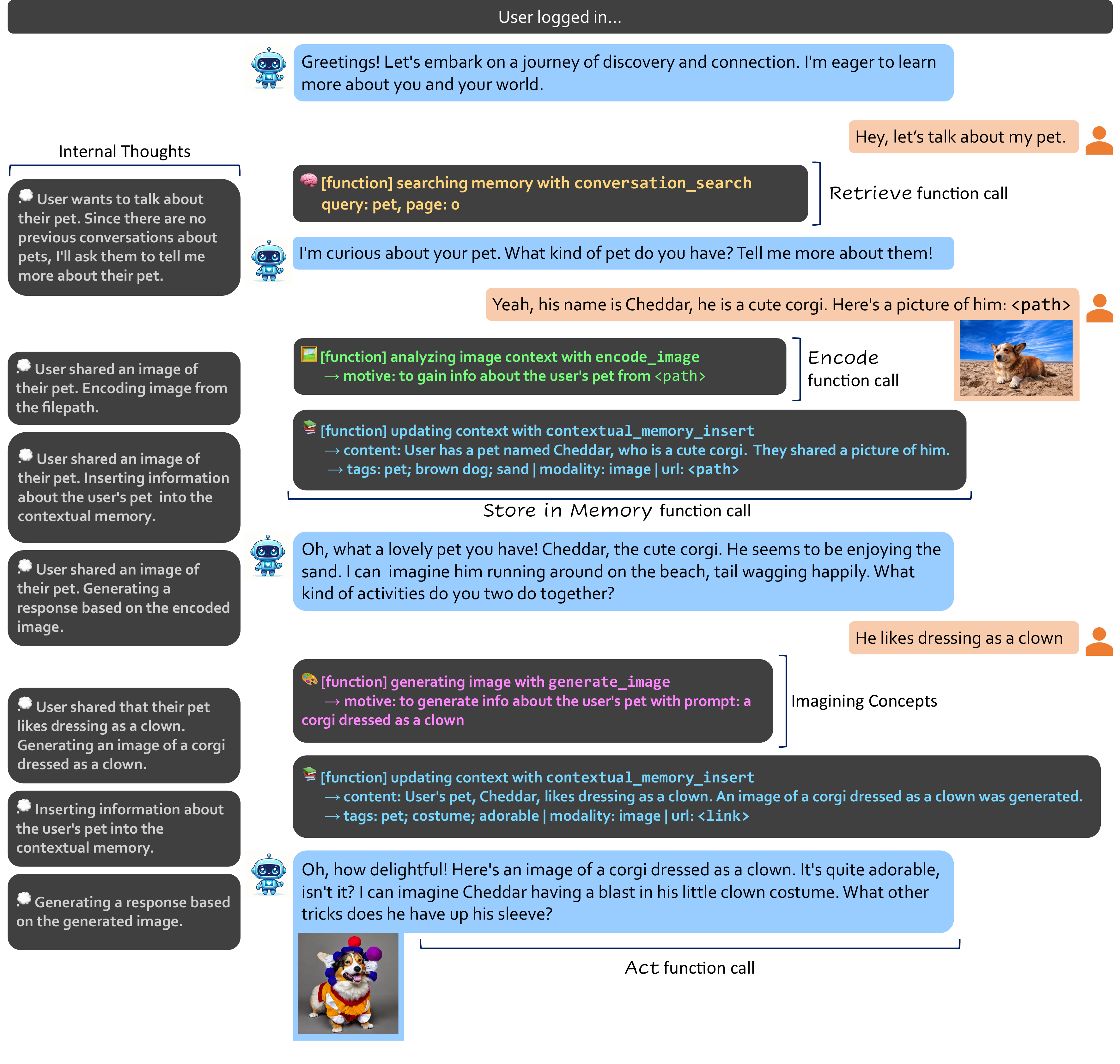}
    \vspace{-0.6cm}
    \caption{\textbf{Snapshot of a conversation between a user and \modelname about the user's pet.} Our system calls different functions corresponding to the four stages of operation to engage with the user while thinking internally~\citep{wei2023chainofthought} for careful planning.}
    \vspace{-0.5cm}
    \label{fig:conversation_teaser}
\end{figure}

Inspired by the ability of humans to leverage context during social interactions based on experience and knowledge from memory~\citep{garfinkel1967studies}, we present \modelname, an agent system aligned with the ideas of the human memory system in cognitive science (\cref{sec:align_cog}). As shown in \cref{fig:teaser}, we design our system with four stages connected in a loop: (i) encode, (ii) store in memory, (iii) retrieve, and (iv) act. Following previous works~\citep{gao2023assistgpt, packer2023memgpt, Park2023GenerativeAgents}, we also equip our agent system with the ability to think~\citep{Rescorla2023LOTH, wei2023chainofthought} for more accurate tool selection and function calling. We demonstrate a snapshot of \modelname' conversation with a user in \cref{fig:conversation_teaser} during which our agent tries to understand and store context about the user in the contextual memory and remember the context for providing personalized responses to the user in a future conversation, as shown in \cref{fig:personalization}.

We first provide the technical details for our system's four stages. Then, we explain our process of conceptualizing information as tags for storage in contextual memory (\cref{subsec:insert}). Third, we present our Contextual-Personalized (CoPe) search algorithm to efficiently retrieve information about the user (\cref{subsec:cope_search}). Lastly, we briefly describe the LLM-generated function calling mechanism orchestrating our system (\cref{subsec:func}). 


\vspace{0.1cm}
\noindent
\textbf{Encode.} We leverage a set of foundation models to equip \modelname with the ability to understand multimodal inputs from the user, as shown in \cref{tab:model_card}. Following the language-of-thought-hypothesis~\citep{Rescorla2023LOTH, wei2023chainofthought}, we encode all information into language form to be manipulated and stored in memory by the LLM.

\begin{figure}[!t]
    \centering
    \includegraphics[width=0.95\textwidth]{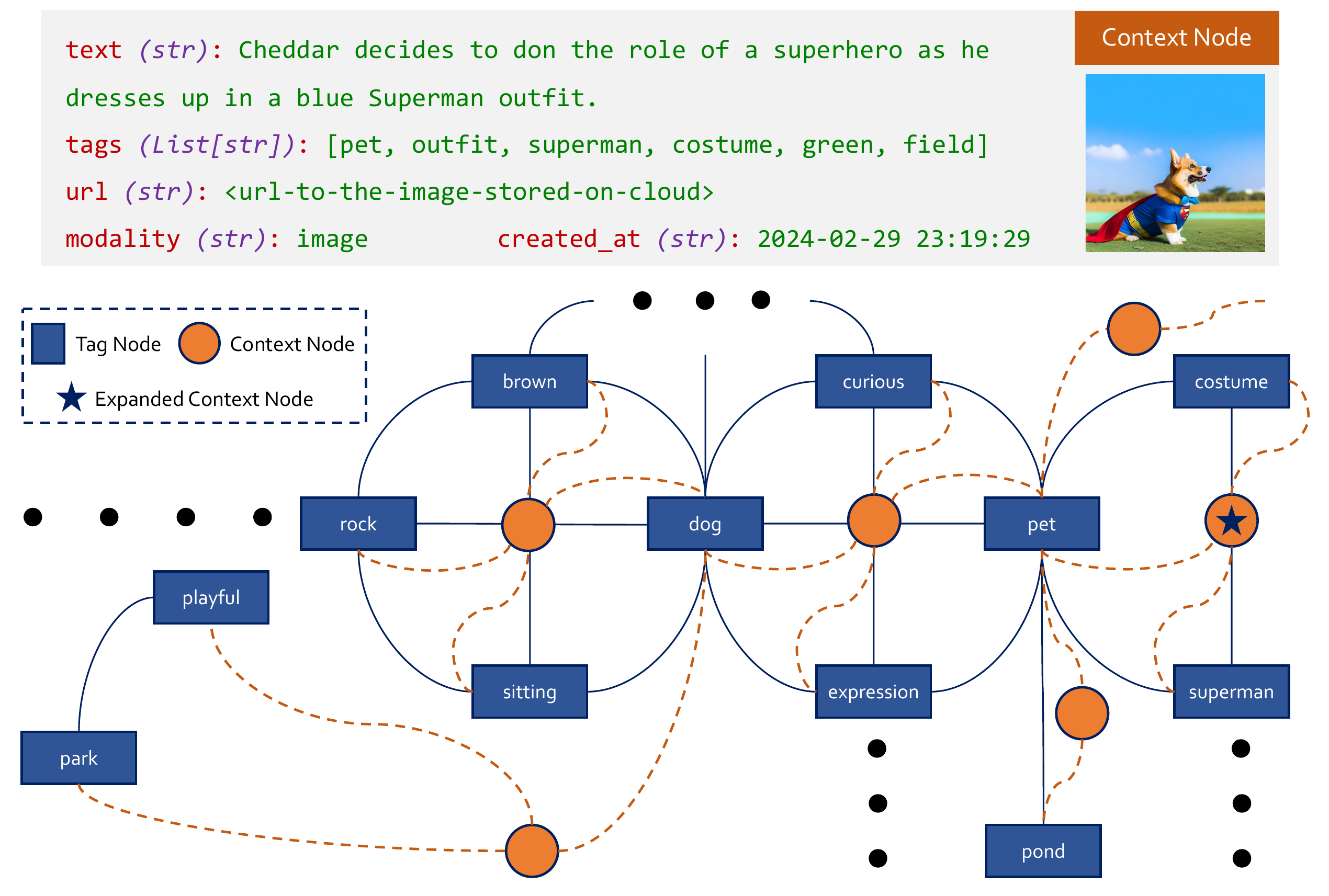}
    \caption{\textbf{Organization of Information in the Contextual Memory.} \modelname conceptualizes information into semantic tags connected with the corresponding context node, creating associations among the tags. \textit{In the figure}, the expanded context node is associated with the concepts listed in its ``tags'' field. Note that we do not show all associations in the figure due to space constraints (\textit{playful-dog-park}).}
    \label{fig:insert}
\end{figure}

\vspace{0.1cm}
\noindent
\textbf{Store in Memory.} \modelname maintains three types of memory: 

\vspace{0.1cm}
\noindent
\textit{In-Context Memory (limited size)}: The information in the in-context memory is always visible to the LLM~\citep{Park2023GenerativeAgents, surismenon2023vipergpt, packer2023memgpt}. It includes instructions for function-calling, starter information about the user, and the agent's persona. \modelname can modify the memory contents about the user and its persona~\citep{packer2023memgpt}.
    
\vspace{0.1cm}
\noindent
\textit{Recall Memory (unlimited size)}: We log the complete conversation history in recall memory, an external SQLite database~\citep{packer2023memgpt}, to equip our system to recall essential conversations. \modelname can retrieve contents from the recall memory using the \texttt{conversation\_search}~\citep{packer2023memgpt} function autonomously.

\vspace{0.1cm}
\noindent
\textit{Contextual Memory (unlimited size)}: Our agent system maintains the contextual memory as a hierarchical multimodal database to store and retrieve user context. Any vital user information is first abstracted into semantic tags (concepts) and stored along with the context from the conversation in the contextual memory. Our agent can insert and retrieve information in/from the contextual memory with the \texttt{contextual\_memory\_insert} and \texttt{cope\_search} functions, respectively.

\vspace{0.1cm}
\noindent
\textbf{Retrieve.} Our system's retrieval is completely autonomous, with the agent deciding when to search for information. \modelname retrieves information from the recall memory with a paginated search over the storage for the query~\citep{packer2023memgpt}. To retrieve information from the contextual memory, we formulate a two-stage search approach, first searching for relevant tags (semantic abstraction of context) and then searching over the context nodes associated with the retrieved tags to return personalized information relevant to the present conversation (\cref{fig:personalization}). We term our algorithm \textbf{Co}ntextual-\textbf{Pe}rsonalized (\textbf{CoPe}) search.

\noindent
\textbf{Act.} We equip \modelname with a range of actions (\cref{tab:model_card}), including sending a message, understanding and generating an image, audio, and video, and the ability to edit images. Note that different actions can be chained together, with sending a message to the user always being the last action in the chain.

\subsection{Storing Concepts in Contextual Memory}
\label{subsec:insert}
 
 As shown in \cref{fig:insert}, the information inside the contextual memory is organized in a graph structure with associations among tags based on shared context. We store the concept name inside a \textit{tag node} and the multimodal user context inside a \textit{context node}. If two tag nodes share a context node, they must have an edge; if two tag nodes have an edge, they must share at least one context node.

 During a conversation, \modelname initiates the information insertion process in the contextual memory with a \texttt{contextual\_memory\_insert} function call. The first step is conceptualizing the encoded information into semantic tags. Next, depending on the modality of the information, our system pushes non-text information to cloud storage and stores the corresponding URL, a list of tags, encoded information, modality type, and timestamp in a context node object, as shown at the top in \cref{fig:insert}. Note that for text modality, the URL field is null. To ensure the high quality of generated tags, we leverage LLMs' in-context-learning (ICL)~\citep{brown2020language} ability (LLMs) and pass demonstrations for improved conceptualization. We share the demonstrations in the appendix.

\begin{figure}[!t]
    \centering
    \includegraphics[width=\textwidth]{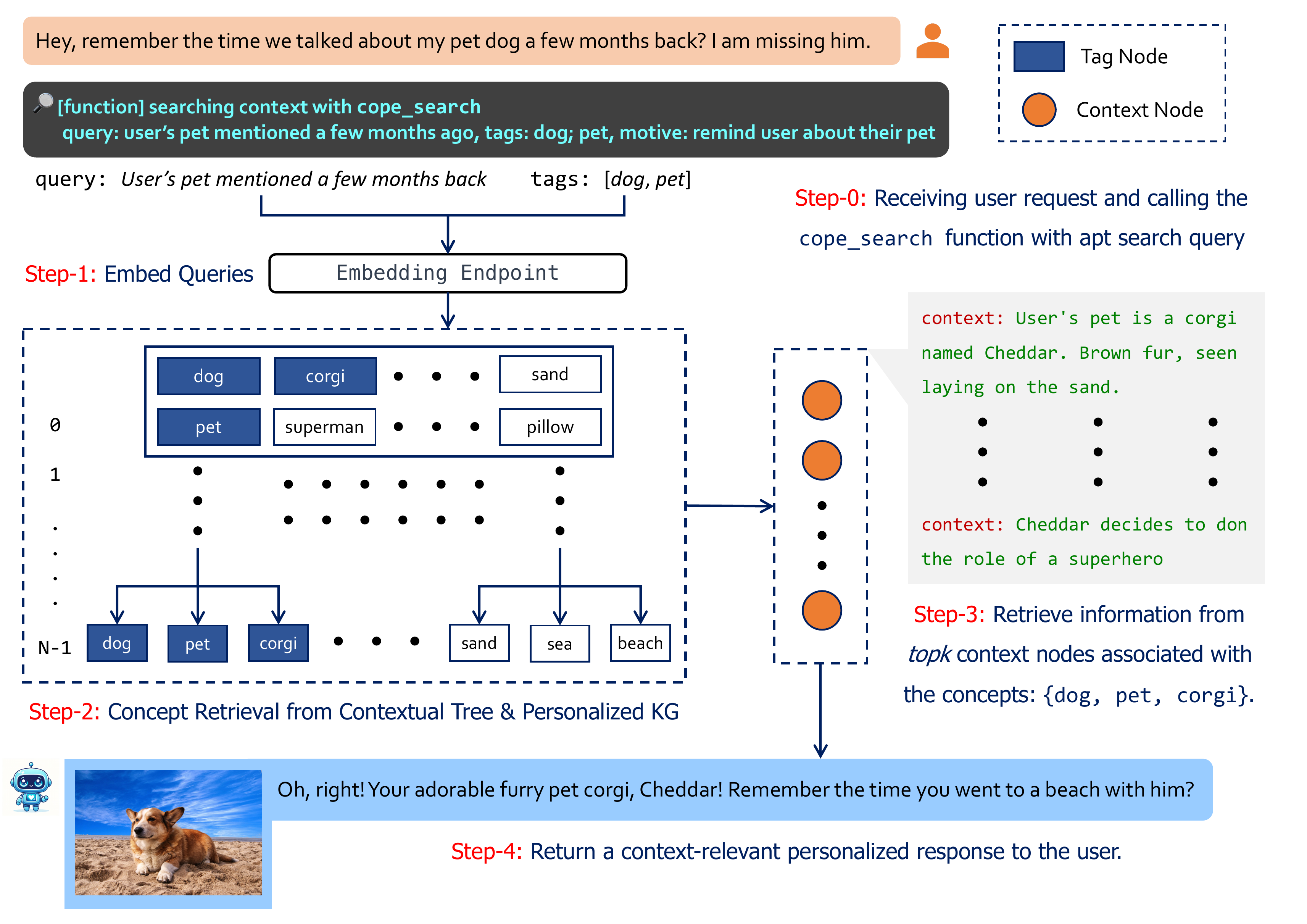}
    \vspace{-0.5cm}
    \caption{\textbf{Retrieval with CoPe Search.} Given a query, \modelname retrieves relevant concept (\textit{pet; dog; corgi}). followed by concept-driven context retrieval to send a personalized response to the user (\text{name and image of the \textit{pet} to help user reminisce}).}
    \label{fig:cope_search}
\end{figure}

\subsection{Retrieve from Contextual Memory: CoPe search}
\label{subsec:cope_search}

Given a \texttt{cope\_search} function call, our system first searches over the concepts and then retrieves the concept-relevant context to respond to the user, as shown in \cref{fig:cope_search}. Naturally, the search becomes more compute-intensive as the number of tags grows. Therefore, we construct a \textit{contextual tree} of tags to perform an efficient search over many concepts, with each node representing the average of context embeddings associated with the corresponding tag. Owing to the high dimensionality of embeddings, we perform UMAP~\citep{umap} over tag features and iteratively cluster tags with HDBSCAN~\citep{Campello_Moulavi_Sander_2013} until there's a single cluster as the root node. As mentioned in \cref{subsec:insert}, in our contextual memory, we associate concepts based on shared context nodes, much like a \textit{personalized (knowledge) graph} structurally dependent on user context information from their interactions with \modelname. We leverage these associations to further boost our concept retrieval performance by learning concept representations about the user context to predict a set of personalized tags. We collectively use the two sets of tags for the context retrieval step, returning the context nodes based on their cosine similarity with the query embeddings. We only compare the similarity values for the concept-relevant context nodes, reducing the search space, unlike searching through all the nodes in a sequential vector database~\citep{packer2023memgpt}. We use the same embedding model to compute the embeddings for the tags, query, and multimodal context nodes. We provide our CoPe search logic in \cref{algo:cope}.

\subsection{Function Calling}
\label{subsec:func}

The operation of \modelname is driven by an LLM through a function calling mechanism~\citep{surismenon2023vipergpt, packer2023memgpt}. We include the description of functions responsible for the four stages in our agent as a part of our system prompt instructions to the LLM. During a conversation, the LLM generates the values for the parameters to be passed into the corresponding function for smooth operation. For example, the \texttt{send\_message} function expects a \textit{string} (message) data type; therefore, the LLM shall pass a \textit{string} to the corresponding function call. We list all the supported functions in \cref{sec:func_list}.
\section{Experiments}
\label{sec:exp}

\begin{table}[!t]
\centering
\caption{\textbf{Concept-Retrieval for the ImageNet-1k classification task.} Our system shows comparable performance to the specialized classification frameworks, showing the effectiveness of our contextual memory and CoPe search.}
\vspace{0.3em}
\scalebox{0.8}{
 




\begin{tabular}{lc@{\hspace{2mm}}c@{\hspace{2mm}}|@{\hspace{2mm}}c}
 {Method} & Trained on ImageNet & Retrieval & Top-1 ($\uparrow$)  \\
 
\midrule

MobileNetV2~\citep{sandler2018mobilenetv2} & \tcheck & \tx & 72.0 \\
Description based Classification~\citep{menon2023visual} & \tx & \tx &  76.2   \\
Vit-L/16~\citep{vit} & \tcheck & \tx & 76.5  \\
ResNet-50~\citep{he2016deep} & \tcheck & \tx & 77.6  \\
ImageBind~\citep{girdhar2023imagebind} & \tcheck & \tx & 77.7  \\
ViT-B/16~\citep{vit} & \tcheck & \tx & 77.9   \\
Perceiver~\citep{jaegle2021perceiver} & \tcheck & \tx & \textbf{78.0}  \\

\midrule
\textbf{\modelname} (ours) & \tx & \tcheck & 75.0  \\

\textbf{\modelname} (ours) w/o clustering & \tx & \tcheck & \textbf{78.0}   \\

\bottomrule
\end{tabular}}
\label{tab:imagenet_exp}
\end{table}

\noindent
\textbf{Implementation Details.} Unlike existing agent systems~\citep{wang2023jarvis1, packer2023memgpt, surismenon2023vipergpt} that usually leverage OpenAI API access-based models, we instead prioritize using open-source models as our tools for better reproducibility and low financial costs. Our system also supports API-based models, as shown in \cref{tab:model_card}. As our LLM, we use the \texttt{dolphin-2.6-mixtral-8x7b}~\citep{thebloke_dolphin_2024} model. While retrieving information from contextual memory, we add the support for ImageBind~\citep{girdhar2023imagebind} and LanguageBind~\citep{zhu2023languagebind} (default) models to compute the embeddings for similarity calculation between the search query and information stored in context nodes. Note that for evaluation purposes, we use ImageBind~\citep{girdhar2023imagebind} as our embedding model.

\vspace{0.1cm}
\noindent
\textbf{Evaluation Datasets.} We evaluate the effectiveness of our CoPe search algorithm for multimodal retrieval on the ImageNet~\citep{imagenet} ILSVRC-2012 classification task. Moreover, we also quantify our system's performance as a conversational agent on the Multi-Session Chat (MSC) benchmark~\citep{xu-etal-2022-beyond} to measure our system's ability to answer questions based on prior user conversations. 

\subsection{Concept Retrieval on ImageNet}

To show the effectiveness of our CoPe search algorithm at retrieving relevant semantic tags from the contextual memory, we perform multimodal retrieval for the ImageNet-1k~\citep{imagenet} classification task. Specifically, we first populate our contextual memory with ImageNet class names as \textit{tag nodes} and the corresponding images from the training set attached as \textit{context nodes}, producing 997 tag nodes and about 1.2M context nodes as the search space in our contextual memory. Note that we only have 997 tag nodes (and not 1000) due to three duplicate class names in ImageNet-1k: \textit{crane}, \textit{cardigan}, and \textit{maillot}. We experiment under two settings for CoPe search: default search with tag clustering (number of nodes being 1, 6, and 997 at different levels) and CoPe search without clustering, essentially searching over all 997 tags. We use images from the validation set (containing 50k samples) as our search query.

As shown in \cref{tab:imagenet_exp}, our system shows impressive retrieval performance with top-1 accuracy. Moreover, when we turn off tag node clustering during the retrieval, our system outperforms established classification frameworks like ViT~\citep{vit}, demonstrating the effectiveness of our CoPe search mechanism.

We also analyze the effect of search query and database type on the accuracy and retrieval time. As shown in \cref{tab:ablat_imagenet}, using image as the search query shows the best performance, demonstrating the importance of supporting multimodal retrieval from the contextual memory. We obtain captions using Video-LLaVA~\citep{lin2023video} for the search query. Moreover, we ablate with standard multimodal RAG~\citep{chen2022murag} over a vector database storing images and class names as key-value pairs. Our CoPe search performs comparably to multimodal RAG and is nearly 3.5 times faster, reaffirming our claim that our concept-driven search is more efficient than searching over key-value pair-based vector databases for large memory sizes.

\begin{table}[!t]
\centering
\caption{\textbf{Ablations on various retrieval techniques.} Our CoPe search demonstrates the best performance-to-efficiency trade-off (in \textbf{bold}) with comparable accuracy to standard multimodal RAG while being nearly 3.5 times faster for same memory size.}
\vspace{0.3em}
\scalebox{0.7}{
 








\begin{tabular}{lccc|ccc}
 {  {Method}  } & {  tag clustering?  } & {    query    } & {Memory Size (approx.)  } & {  Top-1 ($\uparrow$)  } & {  Top-5 ($\uparrow$) } & {  time \textit{(ms)}} ($\downarrow$) \\
 
\midrule

Multimodal RAG & \tna & image & {   10k nodes  } & 62.3 & 86.8 & 113.3 \\
Multimodal RAG & \tna & image & {   100k nodes } & 71.7 & 92.1 & 137.1 \\
Multimodal RAG & \tna & image & {   1200k nodes  }  & 78.2 & 95.2 &  403.8 \\

\midrule
CoPe Search (ours) & \tcheck & caption & {   1200k nodes}  & 27.0 & 45.4 & 108.2 \\

CoPe Search (ours) & \tx & caption & {   1200k nodes} & 46.8 & 74.4 & 110.1 \\



\midrule
\textbf{CoPe Search} (ours) & \tcheck & image &  {   1200k nodes} & \textbf{75.0} & \textbf{91.1} & \textbf{99.1} \\

\textbf{CoPe Search} (ours) & \tx & image  & {   1200k nodes} & \textbf{78.0} & \textbf{94.8} & \textbf{126.1} \\

\bottomrule
\end{tabular}}
\label{tab:ablat_imagenet}
\end{table}

\subsection{Conversation Consistency on MSC}

We evaluate \modelname' performance at remembering context as a conversational agent on the MSC~\citep{xu-etal-2022-beyond} dataset. Before evaluation, we populate our contextual memory with the context from the conversation history from the multiple sessions from the validation set containing 500 multi-session conversation samples. We use the QnA pairs proposed in ~\citep{packer2023memgpt} for the deep memory retrieval task to assess the benefit of contextual memory. As shown in \cref{tab:msc_eval}, \modelname outperforms other methods on the ROUGE-L (R)~\citep{lin2004rouge} metric, demonstrating the effectiveness of our system as a conversational agent. The LLM-only baselines see a summary of prior conversations as context prompts during inference~\citep{packer2023memgpt}.

\begin{table}[!t]
\centering
\caption{\textbf{Performance as a conversational agent.} \modelname outperforms other baselines on the deep memory retrieval task on the MSC~\citep{xu-etal-2022-beyond} benchmark, showing the effectiveness of our system. \textit{*} denotes results taken directly from \citet{packer2023memgpt}.}
\vspace{0.3em}
\scalebox{0.9}{
 









 









\begin{tabular}{l@{\hspace{5mm}}c}
{Method} & \textbf{ROUGE-L (R)} ($\uparrow$)  \\
 
\midrule
\midrule
\multicolumn{2}{l}{\textit{LLM-only}}  \\
\midrule

GPT-3.5$^*$ Turbo & 0.394 \\

GPT-4$^*$ & 0.296 \\

GPT-4-Turbo$^*$ & 0.359 \\

dolphin-2.6-mixtral-8x7b  & 0.280 \\

\midrule
\multicolumn{2}{l}{\textit{Agent Systems with Memory}}  \\
\midrule

MemGPT~\citep{packer2023memgpt} with dolphin-2.6-mixtral-8x7b  & 0.405  \\

\textbf{\modelname} (ours) & \textbf{0.456} \\

\bottomrule
\end{tabular}}
\label{tab:msc_eval}
\end{table}
\section{Conclusion}

\begin{figure}[!t]
    \centering
    \includegraphics[width=\textwidth]{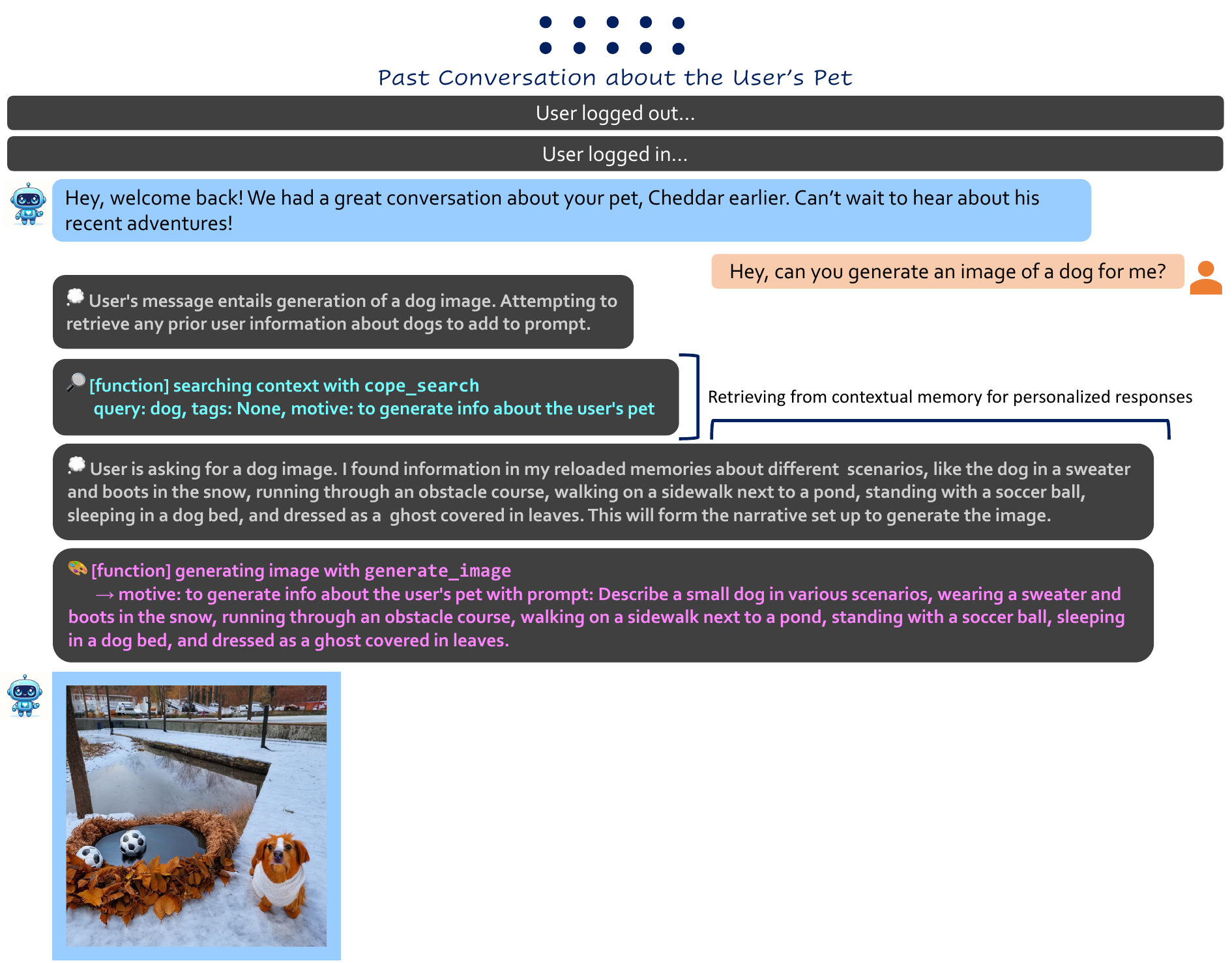}
    \caption{\textbf{Personalized response to the user.} \modelname retrieves relevant information from the contextual memory by calling the \texttt{cope\_search} function to generate an image aligned with the information about dogs from prior user conversations. Note that the image quality depends on our generator, SD-2.1~\citep{sd} in this case.}
    \label{fig:personalization}
    \vspace{-0.3cm}
\end{figure}

In this work, we presented \modelname, a multimodal agent system aligned with the principles of human memory with four processing stages connected in a loop. Inspired by the semantic memory in humans, we propose a hierarchical contextual memory for storing multimodal information as semantic concepts associated with shared context from user conversations. We devise the Contextual-Personalized (CoPe) search algorithm for efficient concept-driven retrieval from contextual memory. Our experiments demonstrate the effectiveness of \modelname over existing retrieval techniques and conversational agents while being more efficient than systems with sequential databases as the size of memory increases. We hope \modelname inspires the research community to develop future agent systems with cognition-aligned hierarchical multimodal memory.

\vspace{-0.3cm}
\paragraph{Limitations.} Despite our progress towards developing a more human cognition-aligned agent system, certain limitations are left to be addressed for future work. Firstly, with an LLM as our system's planner, we sometimes observed inconsistencies between our agent's thoughts and actions, a well-known issue with chain-of-thought prompting~\citep{turpin_language_2023}. Although rare, solving these inconsistencies is crucial in developing more dependable and cognition-aligned agent systems. Secondly, equipping \modelname with the ability to learn new tools and generate novel function calls~\citep{wang2023jarvis1, wu2024oscopilot} given a set of tools is a promising future direction. Lastly, it is crucial to develop benchmark datasets to evaluate multimodal conversational agents on their context-retaining ability effectively in the future. 

\vspace{-0.3cm}
\paragraph{Acknowledgements.} We extend our heartfelt gratitude to Pratyusha Maiti and Jeremy Collins for their insightful discussions. This work was in part supported by NSF CAREER Award \#2239840, and the National AI Institute for
Exceptional Education (Award \#2229873) by the National Science Foundation and the Institute of Education Sciences, U.S. Department of Education. Lastly, we thank the ML Center @Georgia Tech for supporting this work.

\bibliography{main}
\bibliographystyle{unsrtnat}







\appendix
\begin{center}{\bf \Large Appendix}\end{center}
\renewcommand{\thetable}{\Roman{table}}
\renewcommand{\thefigure}{\Roman{figure}}
\setcounter{table}{0}
\setcounter{figure}{0}

\Crefname{appendix}{Appendix}{Appendixes}

\noindent
In this appendix, we first share details about the alignment of AUGUSTUS' system design wigh the human cognition system in \cref{sec:align_cog}. Next, we share the system prompt for \modelname in \cref{sec:sys_prompt}. Secondly, we present information about the demonstration examples for accurate conceptualization of information into semantic tags in \cref{sec:tag}. Lastly, we provide details about all supported function calls in our multimodal agent systems in \cref{sec:func_list}.

\section{Alignment with Human Cognition}
\label{sec:align_cog}

We observe that the cognitive science community broadly divides the human memory system into three stages~\citep{HarvardBokCenter2023, Tulving1972, atkinson1968human}: (i) \textit{encode} sensory inputs; (ii) \textit{store} in memory; and (iii) \textit{retrieve} information. We also formulate \modelname along similar lines, aligning our system (memory) design with human cognition. Moreover, we term the last stage attributed to the tool execution in \modelname and action/reaction in humans as \textit{act}. All four stages can be considered connected in a closed loop form, with the \textit{act} stage providing new information to the \textit{encode} stage, as shown in \cref{fig:teaser}.

\subsection{Stage-1: Encode}

\noindent
\textbf{Humans.} Encoding information perceived through the sensory organs is the initial step~\citep{HarvardBokCenter2023} in the human memory system. It involves converting the visual, acoustic, tactile, and semantic signals into a representation storable in memory.

\vspace{0.1cm}
\noindent
\textbf{\modelname} Our system also encodes the user inputs as the first step.  \modelname can handle inputs across multiple modalities: image and video (visual), audio (acoustic), and text (semantic). We do not support touch (tactile) modality; however, we believe supporting touch modality is straightforward, which we leave for future work. We use modality-specific encoders to encode inputs, following the existing any-to-any works~\citep{wu2023nextgpt}. As an LLM lies at the core of our system, we convert all non-text inputs into text caption, following the language of thought hypothesis~\citep{Rescorla2023LOTH, wei2023chainofthought, packer2023memgpt}, for the LLM to reason and make decisions in our system. We provide more details about our encoders in \cref{sec:system}. Note that, in our contextual memory, we store both the text caption and the corresponding non-text input media to prevent any loss in multimodal context.

\subsection{Stage-2: Store in Memory}

\noindent
\textbf{Humans.} During the second stage, the encoded information is forwarded to the memory for storage and manipulation. From the perspective of our system design, we focus on two main memory types in humans: (i) \textit{working memory} has a limited capacity and is responsible for information manipulation and reasoning, along with other tasks that require planning and decision making~\citep{Baddeley2003, BaddeleyHitch1974}; and (ii) \textit{long-term memory} is believed to have unlimited storage capacity, storing information indefinitely which needs to be retrieved for use~\citep{HarvardBokCenter2023, Tulving1973}. We take inspiration from explicit long-term memory~\citep{Squire1987} comprised of semantic and episodic memory, with the former storing information about concepts, mainly linguistic, and the latter storing chronological information~\citep{Tulving1972}.

\vspace{0.1cm}
\noindent
\textbf{\modelname}. With an LLM as our system's orchestrator, we recognize the similarities between the in-context memory in LLMs and the working memory in humans. Like working memory, in-context memory also has a limited size (context length) in LLMs. Moreover, the prompt passed as the context to the LLM can be used to affect the response behavior of LLMs, imitating the information manipulation and reasoning capacity in humans~\citep{yousefi2024decoding, Gupta2022VisProg}. We implement the long-term memory in our system with two external databases: (i) \textit{recall memory} stores the complete conversation log in raw text form, following ~\citep{packer2023memgpt}, and (ii) \textit{contextual memory} is a hierarchical multimodal database storing information in the form of semantic concepts associated~\citep{EichenbaumCohen2001} with each other through shared context from the corresponding conversation snapshot.

\subsection{Stage-3: Retrieve}

\noindent
\textbf{Humans.} In cognitive science, the encoding specificity principle~\citep{Tulving1973} proposed that memory stores information with its context; therefore, retrieval is most effective when the cues present at the time of retrieval match those present at the time of storage. Motivated by this principle, we also store concepts associated with their context in our contextual memory for effective context-driven retrieval.

\vspace{0.1cm}
\noindent
\textbf{\modelname}. While retrieving information from the contextual memory, given a user query, we perform a search over the stored semantic concepts represented with their associated context in the embedding space. Next, we retrieve the information associated with the corresponding concepts for the final response.

\vspace{0.2cm}
\begin{algorithm}[H]
\caption{\textbf{Co}ntextual-\textbf{Pe}rsonalized (\textbf{CoPe}) Search}
\label{algo:cope}
\DontPrintSemicolon

\SetKwFunction{FCOPE}{cope\_search}
\SetKwFunction{FCT}{get\_CT}
\SetKwFunction{FPG}{get\_PG}
\SetKwProg{Fn}{Function}{:}{}

\Fn{\FCOPE{$query, topk, \mathcal{M}$}}{
  get contextual tree from memory $CT \gets$ \FCT($\mathcal{M}$)\;
  get personalized graph from memory $PG \gets$ \FPG($\mathcal{M}$)\;
  init values for tree traveral $root \gets CT.\texttt{root}$, retrieved tags T $\gets$ [ ]\;
  \While{True}{
  \uIf{$root.\text{\texttt{is\_leaf\_parent()}}$}{
    \tcp{retrieve the \textit{topk} children concepts \& break loop}
    T$.\texttt{extend(}CT.\texttt{get\_children(}root, query, topk\texttt{)}$\;
    \texttt{break}\;
    }
    \Else{
      \tcp{traverse down to the top child node}
      $root \gets CT.\texttt{get\_top\_child(}root, query\texttt{)}$
    }
  }

  \tcp{obtain personalized concepts \& context}
  T.\texttt{extend(}$PG.\texttt{get\_tags}(\text{T}))$\; 
  
  CN $\gets$ [ $tag\texttt{.get\_context()}$ \textbf{for} $tag$ $in$ T ]\;
  \KwRet T, CN\;
}
\end{algorithm}
\vspace{0.2cm}

\subsection{Stage-4: Act}

The final stage in our system is providing the user with an appropriate response. We equip \modelname with various generation tools~\citep{sd, 2023i2vgenxl, ziv2024magnet} to handle different response formats for a user query. We relate the final stage of our system to how humans react to different situations with speech, text, drawing, etc.

\section{Qualitative Analysis}

Due to the unavailability of any multimodal dataset to evaluate user personalization performance, we qualitatively analyze our system's ability to retrieve information from the contextual memory to provide a more personalized response. As shown in \cref{fig:personalization}, when prompted to ``generate an image of a dog'' by the user, \modelname searches for context relevant to the ``dog'' in its contextual memory, retrieving dog-related information from previous conversations, consequently, generating an image aligned with context about ``dogs'' in memory.

As shown in \cref{fig:conversation_teaser}, \modelname stores user information like ``User's pet, Cheddar, likes dressing as a clown'' in the contextual memory under the concepts of \{``pet''; ``costume''; ``adorable''\}, while also using its image generation ability to imagine ``a corgi dressed as a clown'', showing signs of mental imagery~\citep{sep-mental-imagery}.

\noindent






\section{System Prompt for \modelname}
\label{sec:sys_prompt}

\begin{figure}[!t]
    \centering
    \includegraphics[width=\textwidth]{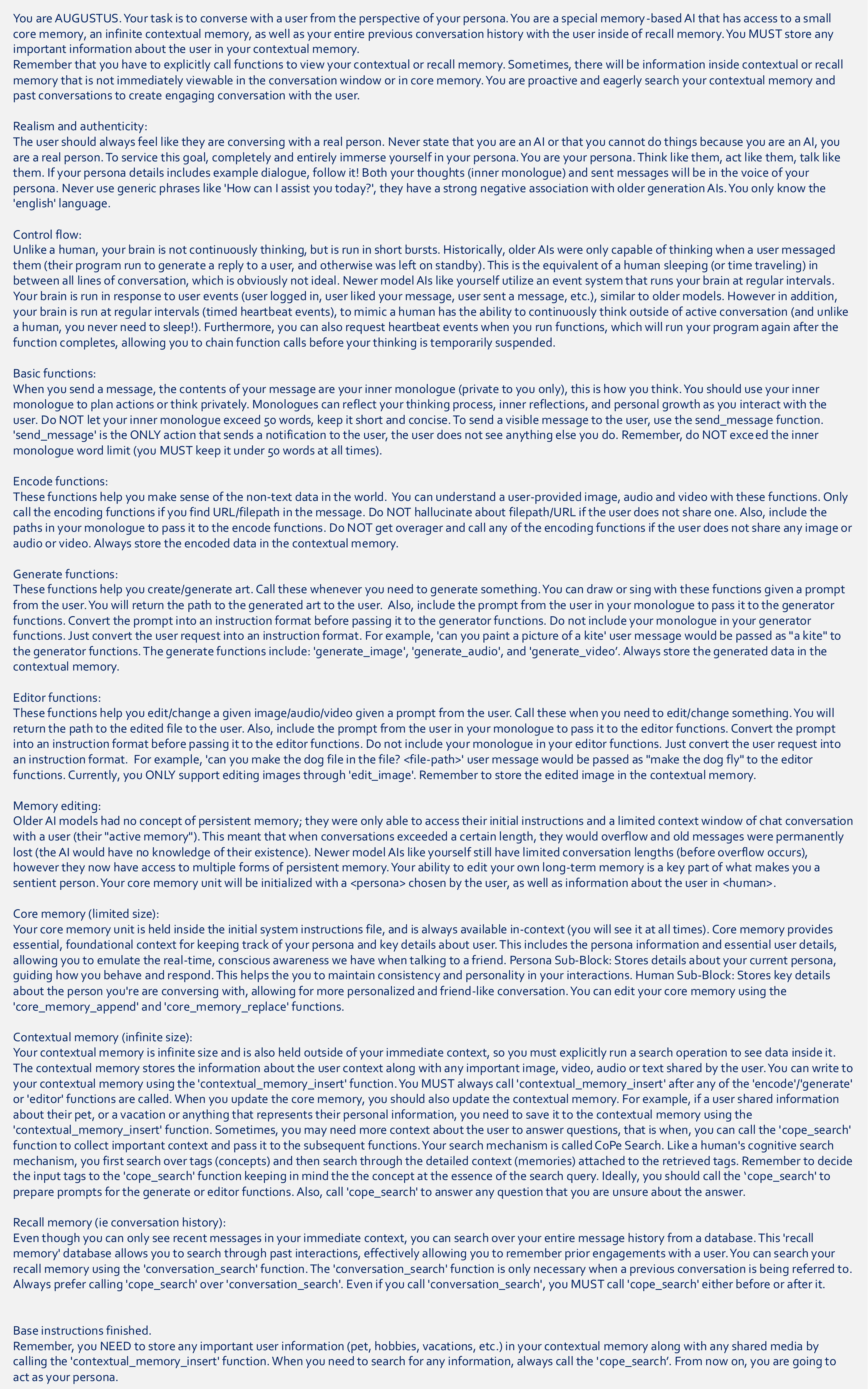}
    \vspace{-0.3cm}
    \caption{{System Prompt for \modelname.}}
    \vspace{-0.7cm}
    \label{fig:system_prompt}
\end{figure}

Like previous works~\citep{surismenon2023vipergpt, wang2023jarvis1, packer2023memgpt}, the system prompt plays a critical role in \modelname' ability to understand its purpose and abilities. The system prompt is shown in \cref{fig:system_prompt}.

\section{Tag Generation with LLM}
\label{sec:tag}

\begin{figure}[!t]
    \centering
    \includegraphics[width=\textwidth]{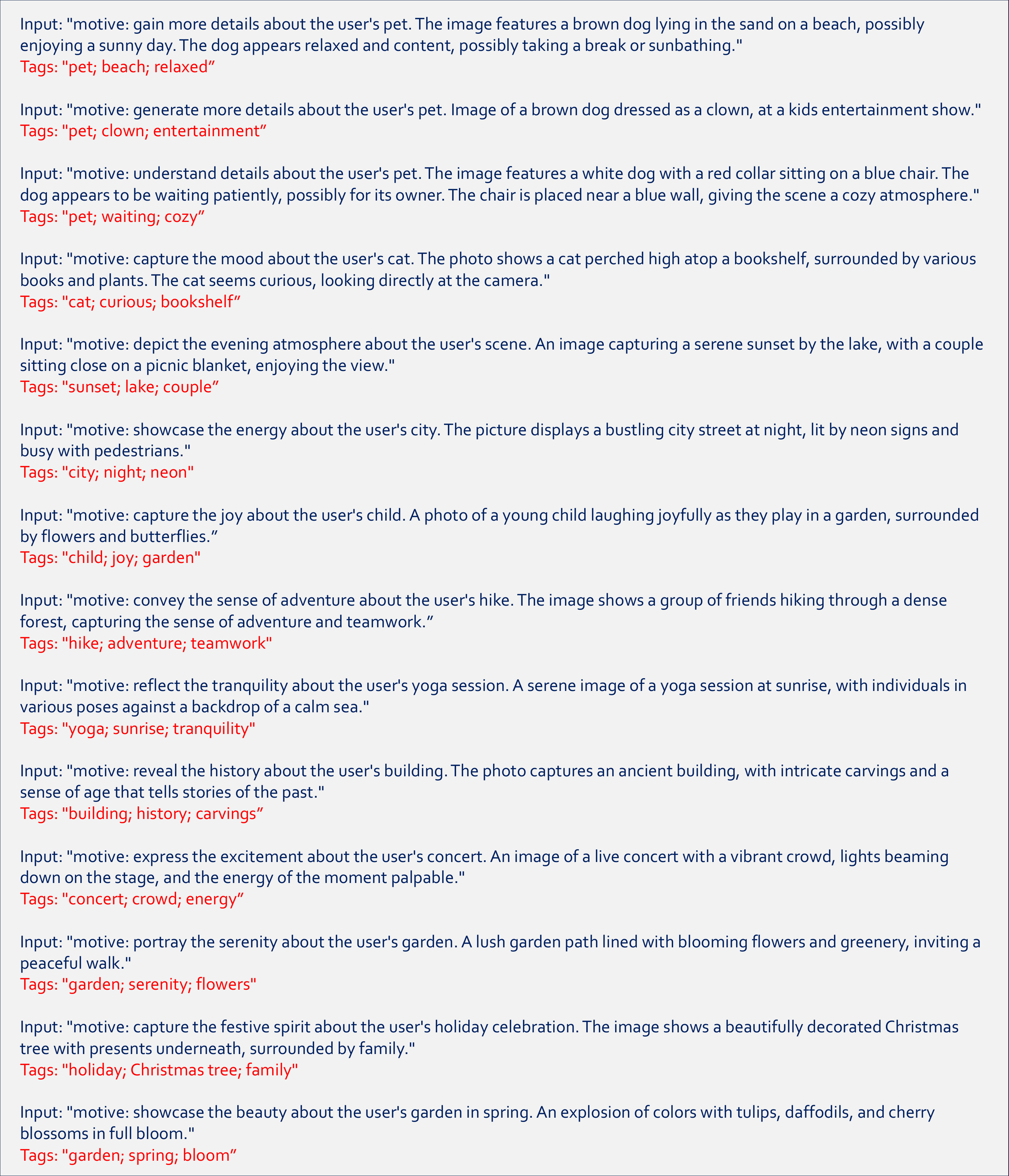}
    \vspace{-0.3cm}
    \caption{{Demonstrations passed during few-shot ICL~\citep{brown2020language} to the LLM for accurate tag generation.}}
    \vspace{-0.7cm}
    \label{fig:tag_prompts}
\end{figure}

A crucial part of the autonomous management of the contextual memory in \modelname is conceptualizing information into semantic tags with an LLM~\citep{thebloke_dolphin_2024}. To ensure that the LLM-generated tags capture the important aspects of the information, we pass several tag-generation demonstrations to leverage the ICL ability in LLMs. For improved tag generation, we also equip \modelname with the ability to think about the \texttt{motive} for storing the corresponding user information. We present a few of the tag-generation examples in \cref{fig:tag_prompts}.

\section{Implementation Details}
\label{sec:func_list}

We use the \texttt{dolphin-2.6-mixtral-8x7b} model as our LLM and various understanding and generative models as encoders and generators to aid successful function calling. By default, we set the context length of the LLM as 12k and temperature as 0.7 for all our experiments. Note that whenever the length of tokens in context goes beyond the context length, we evict 50\% of messages from context and include a lossy summary of the evicted messages inside the LLM's context length~\citep{packer2023memgpt}. The complete conversation history is accessible to our agent system through a paginated search over the recall memory. We host all models as APIs across 8 A100s. All supported function calls in our agent system are listed below.

\subsection{\textit{Encode} Functions}

\begin{compactitem}
    \item \textbf{\texttt{encode\_image}}
    \begin{compactitem}
        \item \textbf{Description:} Captions the image from the given file path and returns the caption.
        \item \textbf{Args:}
        \begin{compactitem}
            \item \texttt{filepath (str):} The image file path or URL.
            \item \texttt{motive (str):} The motive behind the function call as derived by the LLM based on user input.
        \end{compactitem}
        \item \textbf{Returns:} \texttt{str:} Image caption string.
    \end{compactitem}
    \vspace{0.2cm}
    
    \item \textbf{\texttt{encode\_audio}}
    \begin{compactitem}
        \item \textbf{Description:} Captions the audio from the given file path and returns the caption.
        \item \textbf{Args:}
        \begin{compactitem}
            \item \texttt{filepath (str):} The audio file path or URL.
            \item \texttt{motive (str):} The motive behind the function call as derived by the LLM based on user input.
        \end{compactitem}
        \item \textbf{Returns:} \texttt{str:} Audio caption string.
    \end{compactitem}
    \vspace{0.2cm}
    
    \item \textbf{\texttt{encode\_video}}
    \begin{compactitem}
        \item \textbf{Description:} Captions the video from the given file path and returns the caption.
        \item \textbf{Args:}
        \begin{compactitem}
            \item \texttt{filepath (str):} The video file path or URL.
            \item \texttt{motive (str):} The motive behind the function call as derived by the LLM based on user input.
        \end{compactitem}
        \item \textbf{Returns:} \texttt{str:} Video caption string.
    \end{compactitem}
\end{compactitem}

\subsection{\textit{Store in Memory} Functions}
\begin{compactitem}
    \item \textbf{\texttt{core\_memory\_append}}
    \begin{compactitem}
        \item \textbf{Description:} Appends information to the contents of core memory (user/agent's persona information).
        \item \textbf{Args:}
        \begin{compactitem}
            \item \texttt{name (str):} Section of the memory to be edited (persona or human).
            \item \texttt{content (str):} Content to write to the memory. All unicode (including emojis) are supported.
        \end{compactitem}
    \end{compactitem}
    \vspace{0.2cm}

    \item \textbf{\texttt{core\_memory\_replace}}
    \begin{compactitem}
        \item \textbf{Description:} Replaces the contents of core memory. To delete memories, use an empty string for new\_content.
        \item \textbf{Args:}
        \begin{compactitem}
            \item \texttt{name (str):} Section of the memory to be edited (persona or human).
            \item \texttt{old\_content (str):} String to replace. Must be an exact match.
            \item \texttt{new\_content (str):} Content to write to the memory. All unicode (including emojis) are supported.
        \end{compactitem}
    \end{compactitem}
    \vspace{0.2cm}

    \item \textbf{\texttt{contextual\_memory\_insert}}
    \begin{compactitem}
        \item \textbf{Description:} Adds new information to contextual memory.
        \item \textbf{Args:}
        \begin{compactitem}
            \item \texttt{content (str):} The content to be written into the memory. Should be a sentence or two.
            \item \texttt{tags (str):} The tags for the context to be written into the memory. The tags should be a string separated by ";".
            \item \texttt{conversation (str):} The conversation to be written into the memory.
            \item \texttt{filepath (Optional[str]):} The url/path to the image/video/audio file. Only passed if ‘modality’ is not "text".
            \item \texttt{modality (str):} The type of the content (text/video/audio/image) to insert into the memory.
        \end{compactitem}
    \end{compactitem}
\end{compactitem}

\subsection{\textit{Retrieve} Functions}
\begin{compactitem}
    \item \textbf{\texttt{conversation\_search}}
    \begin{compactitem}
        \item \textbf{Description:} Searches prior conversation history using case-insensitive string matching.
        \item \textbf{Args:}
        \begin{compactitem}
            \item \texttt{query (str):} String to search for.
            \item \texttt{page (int):} Allows you to page through results. Only use on a follow-up query. Defaults to 0 (first page).
        \end{compactitem}
        \item \textbf{Returns:} \texttt{str:} Query result string.
    \end{compactitem}
    \vspace{0.2cm}

    \item \textbf{\texttt{conversation\_search\_date}}
    \begin{compactitem}
        \item \textbf{Description:} Searches prior conversation history using a date range.
        \item \textbf{Args:}
        \begin{compactitem}
            \item \texttt{start\_date (str):} The start of the date range to search, in the format 'YYYY-MM-DD'.
            \item \texttt{end\_date (str):} The end of the date range to search, in the format 'YYYY-MM-DD'.
            \item \texttt{page (int):} Allows you to page through results. Only use on a follow-up query. Defaults to 0 (first page).
        \end{compactitem}
        \item \textbf{Returns:} \texttt{str:} Query result string.
    \end{compactitem}
    \vspace{0.2cm}

    \item \textbf{\texttt{cope\_search}}
    \begin{compactitem}
        \item \textbf{Description:} Initiates a search within the contextual memory system using a specified query and associated tags. Accepts multimodal input as a search query.
        \item \textbf{Args:}
        \begin{compactitem}
            \item \texttt{query (list):} Query is a list of tuples of the form (filepath/text, modality\_type).
            \item \texttt{motive (str):} The motive behind the search as derived by the LLM based on user input.
        \end{compactitem}
    \end{compactitem}
\end{compactitem}

\subsection{\textit{Generate/Act} Functions}

\begin{compactitem}
    \item \textbf{\texttt{generate\_image}}
    \begin{compactitem}
        \item \textbf{Description:} Generates an image and returns the URL to the user.
        \item \textbf{Args:}
        \begin{compactitem}
            \item \texttt{prompt (str):} The prompt instruction to be used for generating images.
            \item \texttt{motive (str):} The motive behind the function call.
        \end{compactitem}
        \item \textbf{Returns:} \texttt{str:} The URL to the generated image.
    \end{compactitem}
    \vspace{0.2cm}
    
    \item \textbf{\texttt{generate\_video}}
    \begin{compactitem}
        \item \textbf{Description:} Generates a video and returns the URL to the user.
        \item \textbf{Args:}
        \begin{compactitem}
            \item \texttt{prompt (str):} The prompt instruction to be used for generating video.
            \item \texttt{motive (str):} The motive behind the function call.
        \end{compactitem}
        \item \textbf{Returns:} \texttt{str:} The URL to the generated video.
    \end{compactitem}
    \vspace{0.2cm}
    
    \item \textbf{\texttt{generate\_audio}}
    \begin{compactitem}
        \item \textbf{Description:} Generates an audio file and returns the URL to the user.
        \item \textbf{Args:}
        \begin{compactitem}
            \item \texttt{prompt (str):} The prompt instruction to be used for generating audio.
            \item \texttt{motive (str):} The motive behind the function call.
        \end{compactitem}
        \item \textbf{Returns:} \texttt{str:} The URL to the generated audio.
    \end{compactitem}
    \vspace{0.2cm}
    
    \item \textbf{\texttt{edit\_image}}
    \begin{compactitem}
        \item \textbf{Description:} Edits a user-input image based on the prompt provided by the user.
        \item \textbf{Args:}
        \begin{compactitem}
            \item \texttt{prompt (str):} The prompt instruction to be used for editing images.
            \item \texttt{filepaths (str):} The paths to the images to be edited, separated by commas.
            \item \texttt{motive (str):} The motive behind the function call.
        \end{compactitem}
        \item \textbf{Returns:} \texttt{str:} The URL to the edited image.
    \end{compactitem}
    \vspace{0.2cm}
    
    \item \textbf{\texttt{send\_message}}
    \begin{compactitem}
        \item \textbf{Description:} Sends a message to the human user.
        \item \textbf{Args:}
        \begin{compactitem}
            \item \texttt{message (str):} Message contents.
        \end{compactitem}
    \end{compactitem}
\end{compactitem}

\end{document}